%% file: edlcv.tex
\documentclass[10pt,twocolumn,letterpaper]{article}

\usepackage{cvpr}
\usepackage{times}
\usepackage{epsfig}
\usepackage{graphicx}
\usepackage{amsmath}
\usepackage{amssymb}
\usepackage{subfigure}
\usepackage{multirow}
\usepackage{url}
\usepackage{algorithm,algpseudocode}

\usepackage[pagebackref=true,breaklinks=true,letterpaper=true,colorlinks,bookmarks=false]{hyperref}

\cvprfinalcopy 

\def\cvprPaperID{18} 

\ifcvprfinal\fi
\begin{document}

\title{Fast Hardware-aware Neural Architecture Search}

\newcommand*{\affaddr}[1]{#1} 
\newcommand*{\affmark}[1][*]{\textsuperscript{#1}}
\newcommand*{\email}[1]{\texttt{#1}}

\author{%
	Li Lyna Zhang\affmark[1]
	Yuqing Yang\affmark[1]
	Yuhang Jiang\affmark[2]
	Wenwu Zhu\affmark[2]
	Yunxin Liu\affmark[1]
	\\
	\affaddr{\affmark[1]Microsoft Research Asia},	\affaddr{\affmark[2]Tsinghua University}\\
	\email{\{lzhani, yuqing.yang, yunxin.liu\}@microsoft.com}\\
	\email{\{jyh17\}@mails.tsinghua.edu.cn}, \email{\{wwzhu\}@tsinghua.edu.cn}\\
}

\newcommand{\sysname}{HURRICANE}
\newcommand{\singlepathname}{SPOS}
\newcommand{\algname}{Layerwise-OneShot}
\newcommand{\lz}[1]{{\textcolor{red}{\it LZ: #1}}}
\newcommand{\yyq}[1]{{\textcolor{red}{\it YYQ: #1}}}

\newcommand{\yunxin}[1]{{$\langle${\color{blue} Yunxin: {#1}}$\rangle$}}

\maketitle

\input{abstract}
\input{introduction}

\input{background}

\input{methods}
\input{eval}

\input{conclusion}

{\small
\bibliographystyle{ieee_fullname}
\bibliography{ref}
}
\appendix
\input{models}

\end{document}

%% file: abstract.tex
\begin{abstract}

Designing accurate and efficient convolutional neural architectures for vast amount of hardware is challenging because hardware designs are complex and diverse. This paper addresses the hardware diversity challenge in Neural Architecture Search (NAS). Unlike previous approaches that apply search algorithms on a small, human-designed search space without considering hardware diversity, we propose {\sysname}
that explores the automatic hardware-aware search over a much larger search space and {a two-stage search algorithm}, to efficiently generate tailored models for different types of hardware. Extensive experiments on ImageNet demonstrate 
 that our algorithm outperforms state-of-the-art hardware-aware NAS methods under the same latency constraint on three types of hardware. Moreover, the discovered architectures achieve much lower latency and higher accuracy than current state-of-the-art efficient models. Remarkably, {\sysname} achieves a 76.67\% top-1 accuracy on ImageNet with a inference latency of only 16.5 ms for DSP, which is a 3.47\% higher accuracy and a 6.35$\times$ inference speedup than FBNet-iPhoneX, respectively. For VPU, we achieve a 0.53\% higher top-1 accuracy than Proxyless-mobile with a 1.49$\times$ speedup. Even for well-studied mobile CPU, we achieve a 1.63\% higher top-1  accuracy than FBNet-iPhoneX with a comparable inference latency. 
{\sysname} also reduces the training time by 30.4\% 
compared to {\singlepathname}.


\end{abstract}

%% file: introduction.tex
\section{Introduction}
\label{sec:intro}


\begin{figure*}[t]
	\centering
	\subfigure[Latency and FLOPs on different hardware]{
		\includegraphics[width=0.9\columnwidth]{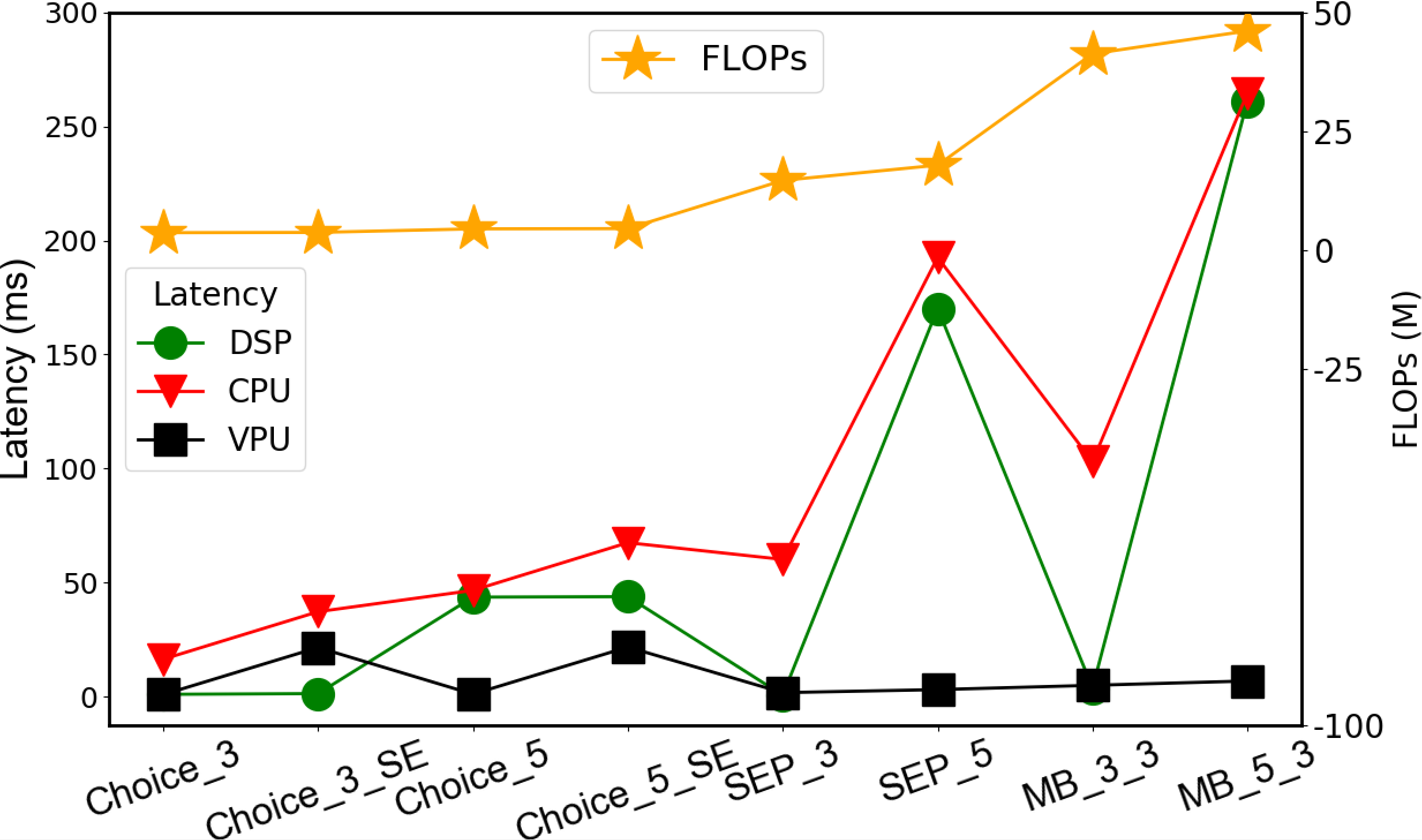}
		\label{fig:op_latency}
		
	}
	\subfigure[Latency in different feature map sizes on VPU]{
		\includegraphics[width=0.84\columnwidth]{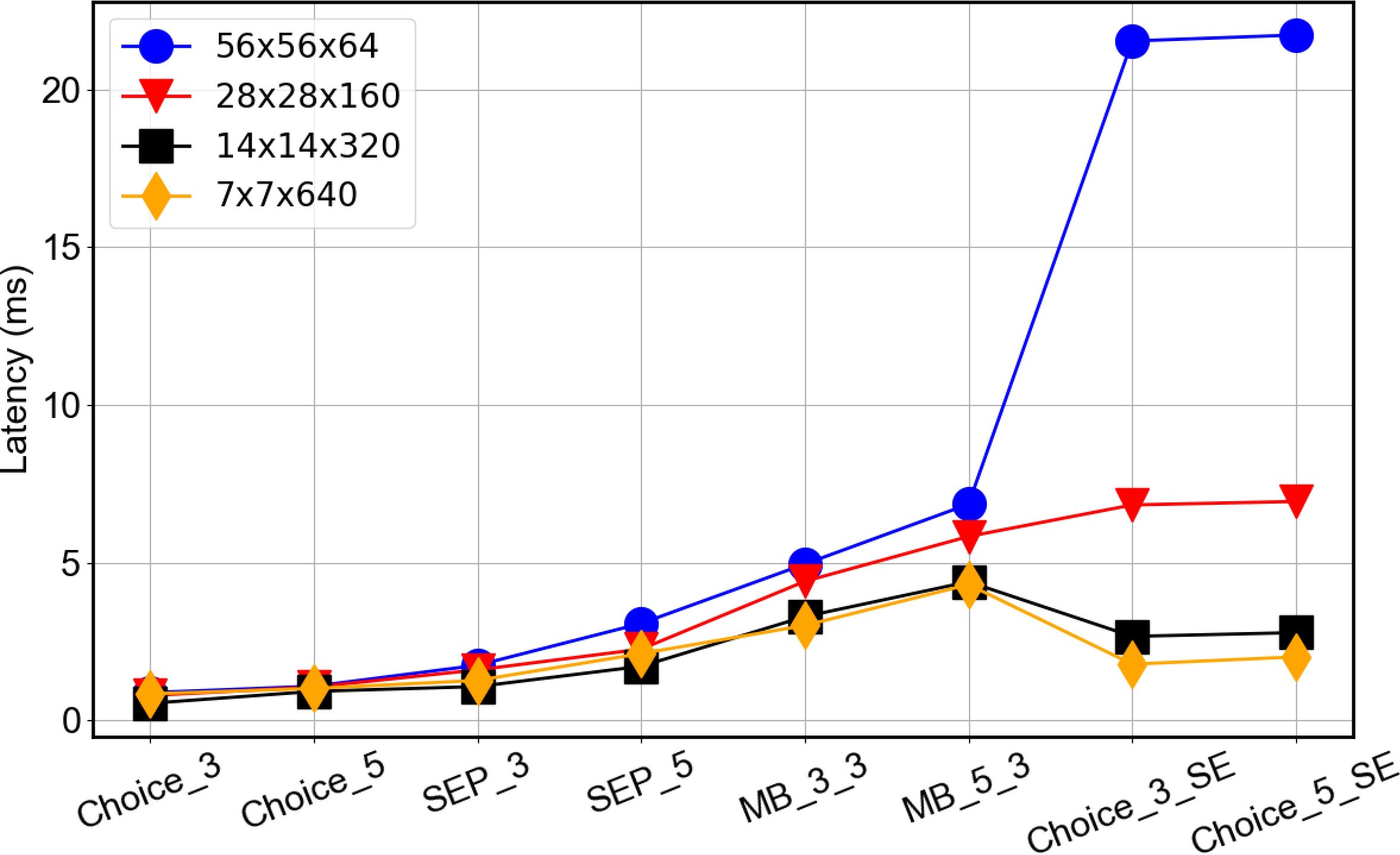}
		\label{fig:op_layer}
		
	}
	
	\caption{Performance of widely used operators in NAS (c.f. Table~\ref{tbl:searchspace}). \textbf{\textit{(a)}}:  Latency and FLOPs on three types of hardware: (1) DSP ($\text{Hexagon}^\text{TM}$ 685 DSP),  (2) CPU (Snapdragon 845 ARM CPU), (3) VPU ($\text{Movidius}^\text{TM}$ $\text{Myriad}^\text{TM}$ X Vision Processing Unit). The input/output feature maps are all the same, equal to $56^2\times64$.  \textbf{\textit{(b)}}: Latency in different input feature map sizes on VPU. }
	\label{fig:motivation}
\end{figure*}

Neural Architecture Search (NAS) is a powerful mechanism to automatically generate efficient Convolutional Neural Networks (CNN) without requiring huge manual efforts of human experts to design good CNN models~\cite{mobilenetv3,rlnas,nasnet,mnasnet, singlepath,oneshot}.  However, searching accurate and fast CNN for the massive smart devices is difficult by current NAS approaches due to the emergence of massive types of hardware devices and the intrinsic huge search cost. 
			

\textbf{Unaware of Hardware Diversity}. Most of previous NAS methods focus on searching for the most accurate models. The common effort to guarantee the inference efficiency (e.g., model inference latency on real hardware) is to limit the model's FLOPs~\footnote{In this paper, the definition of $FLOPs$ follows~\cite{shufflenetv1}, i.e., the number of multiply-adds.}. Some recent hardware aware NAS methods~\cite{singlepath,mnasnet,fairnas,fbnet} consider model-inference performance but they only aim at the same type of hardware, smart phones from different manufacturers but all based on ARM processors. Also, such hardware-aware approaches~\cite{proxyless,fbnet,4hoursnas} use an identical manually elaborated search space for different types of hardware platforms.
However, the emerging massive smart devices (e.g., IoT devices) are equipped with very diverse processors, such as GPU, DSP, FPGA, and various AI accelerators that have fundamentally different hardware designs. Such a big \emph{hardware diversity} makes FLOPs an improper metric to predict model-inference performance and manual-designed search space not ideal for searching efficient models. As a result, it calls for new methods to automatically generate the hardware-aware search spaces that leverage the characteristics of every hardware platform and relax the reliance on human design  effort.

To demonstrate it, we conduct an experiment to measure the performance of a set of widely used neural network operators (a.k.a. operations) on three types of mobile processors: $\text{Hexagon}^\text{TM}$ 685 DSP, Snapdragon 845 ARM CPU, and $\text{Movidius}^\text{TM}$ $\text{Myriad}^\text{TM}$ X Vision Processing Unit (VPU). Figure~\ref{fig:motivation} shows the results and we make the following key observations. First, from Figure~\ref{fig:op_latency}, we can see that even the operators have similar FLOPs, the same operator may have very different inference latency on different processors. For example, the latency of operator \textit{Choice\_3} is almost the same as \textit{Choice\_3\_SE} on the DSP, but the difference on the VPU is more than 24$\times$. Therefore, FLOPs is not the right metric to decide the inference latency on different hardware. Second, the relative effectiveness of different operators on different processors is also different. For example, operator \textit{Choice\_3} has the smallest latency on the ARM CPU, but operator \textit{SEP\_3} has the smallest latency on the DSP. Thus, different processors should choose different operators for the best trade-off between model accuracy and inference latency. Furthermore, as shown in Figure~\ref{fig:op_layer}, the computational complexity and latency of the same operator are also affected by the execution \textit{context}, such as input feature map shapes, number of channels, etc. Such a context is determined by which layer the operator is placed on. That is, even for the same type of hardware, optimal operators may change at different layers of the network. Thus, it is difficult to cover hardware diversity using manually-designed search space.

Motivated by these observations, we argue that there is no one-size-fits-all model for different hardware platforms, and thus propose {\sysname} (shown in Figure~\ref{fig:nascompare}) to generate different models tailored for different types of hardware. 
To cover the diversity of hardware platforms, we construct a much larger candidate operators pool (32 in our implementation) and propose a {search space generation approach} to
automatically generate a \textit{hardware-specialized search space} for each type of hardware. The key point here is to include much more hardware-efficient and accurate candidate architectures in search space without increasing the search cost.  Our mechanism is based on profiled real performance on the target hardware as shown in Sec~\ref{sec:searchspace}.


Moreover, we propose a {two-stage search algorithm} for the one-shot NAS\footnote{In this paper we adopt one-shot NAS because of its simplicity, however, our acceleration could also be combined with other NAS methods.} to further reduce the intensive search cost.  Unlike previous works that search operators for all layers at one time, we search the complete architecture by a sequence of simpler searching of sub-networks. The method is inspired by the \textit{layer diversity} (different layers have different impacts on inference latency~\cite{mobilenetv3} and model accuracy~\cite{unequal_layer,cnnprune}), we demonstrate that exploring more architecture selections in the  layers close to classification output may help find better architectures with the limited sampling budget, and limiting the latency in the layers close to data input is critical to search for low-latency models. 

In summary, we propose a novel approach that enables NAS to quickly search the accurate and fast architectures for different type of hardware platforms (not only the ARM CPU). The key technical innovations are: (i) automatically generate more effective search space for target hardware with minimal human design, (ii) explore more architectures in deeper  layers and reduce search space size.  We conduct comprehensive experiments on ImageNet 2012 and OUI-Adience-Age datasets over three hardware platforms (Figure~\ref{fig:op_latency}). Under all the three platforms, {\sysname} consistently achieves the  better accuracy than state-of-the-art hardware-aware NAS methods with the same latency constraints. Specifically,  {\sysname} improves the top-1 ImageNet accuracy by an average of 1.83\% than Proxyless~\cite{proxyless}, and 1.35\% than SPOS~\cite{singlepath}.  In addition, the searched architectures also outperform the current state-of-the-art efficient models. Remarkably, {\sysname} reduces the  latency by 6.35$\times$ on DSP compared to FBNet-iPhoneX and 1.49$\times$ on VPU compared to Proxyless-mobile, respectively. Finally, {\sysname} reduces the training time by 30.4\% on ImageNet comparing to SPOS.



%% file: background.tex
\section{Related Work}

\textbf{Hardware aware NAS.}  
 Recent methods~\cite{mnasnet, fbnet,proxyless,singlepath,oneshot} adopt a layer-level hierarchical search space with a fixed macro-structure allowing different layer structures at different resolution blocks of a network. The goal becomes searching operators for each layer so that the architecture achieves competitive accuracy under given constraints. To search hardware-efficient architectures, the search spaces have been built on increasingly more efficient building blocks. ~\cite{mnasnet,proxyless,4hoursnas} built upon the MobileNetV2~\cite{mobilenetv2} structure (\textit{MB\_k\_e}).  \cite{fbnet,singlepath}  built search space  by ShuffleNetV1~\cite{shufflenetv1} and ShuffleNetV2~\cite{shufflenetv2} (\textit{Choice\_k}). As these structures are primarily designed for mobile CPU, the efficiency of such manually-designed search space is unknown for other hardware.
 
  To measure the model efficiency, many NAS methods~\cite{rlnas,nasnet} adopt the hardware-agnostic metric FLOPs. However, architecture with lower FLOPs is not necessarily faster~\cite{hyperpower}. Recently, gradient-based methods~\cite{proxyless,4hoursnas,fbnet} adopt direct metrics such as measured latency but only for mobile CPUs. They profile every operator's latency and build prediction model. The latency is then viewed as a differentiable regularization loss. However, the multi-objective loss is not optimal because accuracy changes much more dramatically with latency for small models, as ~\cite{mobilenetv3} pointed out. Instead, we follow One-Shot NAS~\cite{singlepath,fairnas} and apply the latency constraints directly.


\textbf{One-Shot NAS.} Starting from ENAS~\cite{enas}, weight sharing became popular as it accelerates the search process and makes search cost feasible. Recent one-shot methods encode the search space into an over-parameterized supernet, where  each path is a stand-alone model. During the supernet training, architectures are sampled  by different proxies (e.g., reinforcement learning) with weights updated. However, {\singlepathname}~\cite{singlepath} and FairNAS~\cite{fairnas} observe that such coupled architecture search and weight sharing could be problematic to fairly evaluate the performance of candidate architectures. SPOS~\cite{singlepath} trains the supernet by an uniform path sampling method, and applies an evolutionary algorithm to efficiently search architectures directly without any fine tuning. As it's easy to train and fast to search, our work is built upon  {\singlepathname}~\cite{singlepath} with their officially open-sourced implementation~\cite{singlepath_code}.

%% file: methods.tex
\section{Methodology}
\label{sec:method}

\begin{figure}[t]
	\begin{center}
		\includegraphics[width=1\columnwidth]{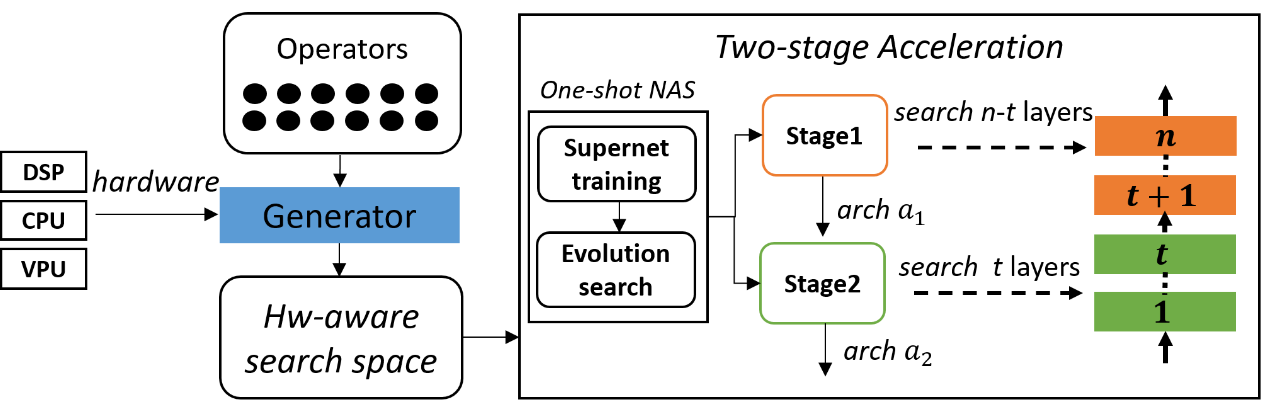}
		%
			
	\end{center}
	\caption{
		{\sysname} constructs hardware-specialized search space (by latency constraints) that contains more efficient architectures than previous NAS, and employs a two-stage search algorithm to reduce the search cost. }
	\label{fig:nascompare}
\end{figure}

\begin{table*}[h]
	\begin{center}
		\small
		\begin{tabular}{|c|c|c|c|c|c|}
			\hline
			
			\textbf{Output shape} & \textbf{Layer} & {\textbf{DSP}} &  {\textbf{CPU}} &  {\textbf{VPU}} \\
			\hline
			\multirow{ 2}{*}{$56^2\times64$}&\multirow{ 2}{*}{1-4} & SEP\_3, Choice\_3 & Choice\_3, Choice\_3\_SE              &   Choice\_3, Choice\_5                \\
			&&  MB\_3\_1, ChoiceX  &  MB\_3\_1, ChoiceX        &      Choice\_7, SEP\_3           \\
			\hline
			\multirow{ 2}{*}{$28^2\times160$}&\multirow{ 2}{*}{5-8} & Choice\_3, ChoiceX  & Choice\_3, ChoiceX & Choice\_3, Choice\_5 \\
			& & MB\_3\_1, Choice\_3\_SE  &Choice\_5, MB\_3\_1  &Choice\_7, ChoiceX \\
			\hline
			\multirow{ 2}{*}{$14^2\times320$} &\multirow{ 2}{*}{9-16}& Choice\_3, Choice\_3\_SE & Choice\_3, Choice\_3\_SE  & Choice\_3, Choice\_5 \\
			&&ChoiceX, MB\_3\_1& Choice\_5 ,Choice\_5\_SE &Choice\_7, ChoiceX \\
			\hline
			\multirow{ 2}{*}{$7^2\times640$}&\multirow{ 2}{*}{17-20}& Choice\_3, Choice\_3\_SE &Choice\_3, Choice\_5  & Choice\_3, Choice\_5\\
			&&ChoiceX  ,MB\_3\_1, MB\_3\_3 &Choice\_3\_SE, Choice\_7, MB\_5\_1  &Choice\_7, MB\_3\_1, MB\_7\_1 \\
			\hline
			
		\end{tabular}
	\end{center}
	\caption{Hardware-aware search space for each mobile hardware. For  layer at 1-16, it contains 4 operators for selection, for layer 17-20, each layer has 5 operators. The input/output channel and stride settings for each layer are the same with {\singlepathname}~\cite{singlepath}. }
	\label{tbl:searchspace}
\end{table*}

In this paper, {\sysname} aims to search the following architectures for a given hardware platform $h$ (any of CPU, DSP, NPU, VPU, etc.) and the latency constant $\tau_c^{(h)}$:

\begin{equation}
	\begin{aligned}
		& \displaystyle \max \quad \text{ACC}_\text{val}(a) \\
		& \displaystyle \textrm{s.t.} \quad \tau(a,h) \leq \tau_c^{(h)}
	\end{aligned}
\label{eq:problem}
\end{equation}
We seek to find an architecture $a$ that achieves the maximum accuracy ACC$_\text{val}(a)$ on the validation set while the inference latency $\tau(a,h)$ is under the constraint $\tau_c^{(h)}$. 

\subsection{Hardware-aware Search Space}
\label{sec:searchspace}
We follow the design of layer-level hierarchical search space in recent hardware-aware NAS~\cite{fbnet,mnasnet}. Besides first and last three fixed layers, each learnable layer can choose an operator from a candidate pool. For each target hardware, we encode the specialized search space into a over-parameterized supernet for one-shot NAS.

\begin{table}[htbp]
	\begin{center}
		\footnotesize
		\begin{tabular}{|c|c|c|c|}
			\hline
			\multirow{2}[4]{*}{\textbf{Operator}} & \multicolumn{2}{c|}{\textbf{Variable range}} & \multirow{2}[4]{*}{\textbf{Number}} \\
			& \textbf{Kernel ($k$) }    &\textbf{ Expansion ($e$) }    &  \\
			\hline
			SEP\_$k$ & 3,5,7 &   -    & 3 \\
			SEP\_$k$\_SE & 3,5,7 &  -     & 3 \\
			MB\_$k$\_$e$ & 3,5,7 & 1,3,6 & 9 \\
			MB\_$k$\_$e$\_SE & 3,5,7 & 1,3,6 & 9 \\
			Choice\_$k$ & 3,5,7 &   -    & 3 \\
			Choice\_$k$\_SE & 3,5,7 &  -     & 3 \\
			ChoiceX &   3    &  -     & 1 \\
			ChoiceX\_SE &    3   &   -    & 1 \\
			\hline
		\end{tabular}%
	\end{center}
	\caption{Candidate operators. For depthwise convolution in each operator, we  allow choosing $k$ of 3 , 5, 7. For the expansion ratio $e$ in \textit{MB}, we allow it choosing of 1, 3, 6. }
	\label{tbl:operators}%
\end{table}%

\textbf{Diverse Candidate Operators Pool}.
Compared with the small operator pool in previous works, we employ a much bigger pool of candidate operators from the primary blocks of off-the-shelf networks. In our experiment, FLOPs and memory access cost of an operator leverage different impacts to the latency on three hardware platforms. As a result, our pool contains up to 32 operators (detailed in Table~\ref{tbl:operators}) with different levels of  computation and memory complexity. They are built upon the following 4 basic structures from current efficient models:

\begin{itemize}
	\item \textbf{SEP}: depthwise-separable convolution. Following DARTS~\cite{darts}, we applied the depthwise-separable convolution twice. 
	 \textit{SEP} has a larger FLOP count than others, but less  memory access complexity. 
	 \item \textbf{MB}: mobile inverted bottleneck convolution in MobileNetV2~\cite{mobilenetv2}.
	    \textit{MB} has a medium memory access cost due to its shortcut and add operation. Its computation complexity is decided by the kernel size \textit{k} and expansion rate \textit{e}.
	
	\item \textbf{Choice}: basic building block in ShuffleNetV2~\cite{shufflenetv2}. 
	Following~\cite{singlepath}, we add a similar operator \textit{ChoiceX}.  \textit{Choice} and \textit{ChoiceX} have much smaller FLOPs than the others, but the memory complexity is high due to the channel split and concat operation. 
	\item \textbf{SE}:  squeeze-and-excitation network~\cite{se}. To balance the impacts in latency and accuracy, we follow the settings in MobileNetV3~\cite{mobilenetv3}. We set the reduction ratio $r$ to 4, and replace the original sigmoid function with a hard version of swish $hswish[x]=x\frac{ReLU6(x+3)}{6}$. We apply \textit{SE} module to the above operators and generate new operators.  The computation complexity of \textit{SE} is decided by its insert position, while the memory access cost is relatively lower.
\end{itemize}


\textbf{Hardware Aware Search Space Generation}. 
The enrichment of candidate operators covers the diversity of hardware platforms.  However, doing so increases the search space by many
orders of magnitude (e.g., the original large search space size is $10^{18}$ larger than SPOS), and thus leads to unacceptable search and training cost and may even cause non-convergence problem in  one-shot  NAS methods.

To reduce the cost while improving the search space efficiency, we propose a layer wise \textit{hardware aware search space generation} approach to generate specialized search space for \textit{every target hardware platform}.  Unlike the previous methods that apply same operators for all layers, we select the most efficient operators for every layer by real hardware deployment score.  We benchmark all the 32 operators layer-by-layer and sort each layers' candidate operators in non-increasing order of their scores in Equation 2:

\begin{equation}
\label{eq:score}
\displaystyle score^{(i)}_{op} = (F_{op}\times P_{op})^\alpha (\tau_{\ell_i(a)=op}(a,h))^{-1}
\end{equation}
where $F_{op}$ and $P_{op}$ are the FLOPs count and number of parameters of operator $op$ respectively, $\ell_i(a)=op$ means architecture $a$ whose $i$-th learnable layer is $op$. Parameter $\alpha$  is non-negative constant. The score of candidate operator $op$ at the $i$-th learnable layer ($score^{(i)}_{op}$) considers both representation capacity (approximately) and real hardware performance.

 The operators listed upfront will be selected to construct the reduced search space. For each layer, we filter out the first $p$ operators with highest score, and the size of search space would be $n^p$ ($n$ is the number of learnable layers).  In our experiment, we choose the top $p=4$ operators for each layer to keep the similar size with other NAS search space.  

\textbf{Exploring Operator}.
Inspired by the observations of \textit{layer diversity}, some layers (commonly in the layers close to output) contribute small to the latency (due to the small feature map size) but impact largely on the accuracy. For these learnable layers, we add an extra exploring operator besides the first $p$ operators. Since exploring operator is mainly for better accuracy, its score could be not so top ranked. For our backbone network (shown in Table~\ref{tbl:searchspace}), it is natural to add the exploring operator to the last 4 layers because of their smallest feature map size.

In summary, we construct three different search spaces for our hardware platforms as shown in Table~\ref{tbl:searchspace}. For every specialized search space, it contains $n=20$ learnable layers, and each layer can choose from 4 or 5 candidate operators from the Table ~\ref{tbl:searchspace}.  Each search space contains $4^{16}\times5^4 \approx2\times10^{13}$ possible architectures, which is approximately twice the size of {\singlepathname}'s search space. 

\subsection{Two-Stage NAS Acceleration}
To search an architecture of $n$ learnable layers, early NAS methods search for one cell structure and repeat it for all layers~\cite{rlnas,google_evo}, while the recent NAS methods search operators for the complete architecture~\cite{fbnet,proxyless,singlepath,mnasnet}. We adopt a different two-stage approach that each stage searches operators for part of the whole architecture. This strategy leverages the layer diversity in accuracy and latency, and further to reduce the one-shot search cost.



\textbf{Layer Grouping}. The 
phenomenon that different CNN layers reveal different sensitivity to accuracy has been observed in other domains~\cite{vis_cnn, cnnprune,unequal_layer}. NAS should take more efforts in searching the ideal operators for the critical layers as the operator selection for  non-critical layers impacts less to the final accuracy. However, it's difficult to do the accuracy sensitivity analysis for individual layer in NAS scenario.  Fortunately,
some previous works~\cite{vis_cnn,girish2019unsupervised} have revealed different behaviors between the earlier layers (close to data input) and the later layers (close to classification output) in CNN models. The earlier layers extract low-level features from inputs (e.g., edges and colors), are computation intensive and demand more data to converge, while the later layers capture high-level class-specific features but are less computation intensive. Inspired by this, our intuition is that operator search for later layers is more critical than earlier layers.
To this end, we group the $n$ layers of the CNN model into two groups: the earlier $t$ layers (less critical)  and the later $n-t$ layers (more critical). 

\textbf{Two-Stage Search Algorithm.}
Algorithm~\ref{alg:ha-nas-2steps} illustrates the main procedures of hardware-aware NAS with two-stage search acceleration. Each stage starts with a different winning architecture and runs a one-shot NAS to search the target group of learnable layers. We treat the rest group  as the non-active fixed layers and use the corresponding layers' operator of the winning architecture. 
In the beginning, we set up the initial winning architecture $a_{win0}$ with the operators of the highest scores in every layer (line 4-6). 

First, \textit{Stage1}  searches the later $n-t$ layers for $a_{win0}$. We mark the later $n-t$ layers as active and the earlier $t$ layers as non-active. The non-active layers are fixed to the corresponding layer structures of architecture $a_{win0}$ (line 9), while the active layers will be searched from
the generated operator list $l_i(\mathcal{A})$ (line 10-12). The one-shot NAS method itself is similar to the work~\cite{singlepath}, except that we constraint the search space with a hardware latency other than FLOPs (line 14). After a complete process of one-shot search, a new winning architecture $a_{win1}$ would be generated. 

Second, \textit{Stage2} starts with the new winning architecture $a_{win1}$  and searches for earlier $t$ layers. The later $n-t$ layers are fixed to the corresponding layer operator of $a_{win1}$ and the earlier $t$ layers are active for another one-shot search. \textit{Stage2} returns the final architecture $a_{win2}$.

\algrenewcommand{\algorithmiccomment}[1]{$\triangleright$ #1}
\begin{algorithm}[h]
	\small
	\caption{Hardware-aware NAS with acceleration}\label{alg:ha-nas-2steps}
		\textbf{Input}: hardware $h$, latency constraint $\tau_c^{(h)}$, hyper-parameter $t$
	\begin{algorithmic}[1]

		\Function{TwoStageConstrainedNAS}{$A$,  $t$}
			\State \Comment{$\ell_i(a)$ denotes the $i$-th layer of architecture $a$}
			\State \Comment{$\ell_i(A)$ denotes all the candidate operators in the $i$-th layer of search space $A$, sorted in non-increasing order of score}
			\For{$i \gets 1$ to $n$} 
				\State $\ell_i(a_{win}) \gets \ell_i(A)[0]$  \Comment{Init  with top-1 ranked operator}
			\EndFor
			\Statex
			\State \Comment {Each stage searches 1 group layers by one-shot NAS.}
			\For{$iter \gets 1$ to 2} \Comment Stage1: iter=1, Stage2: iter=2
				\State Fixed $\gets $  $[1,t]$ if $iter=1$, else: $[t+1,n]$
				\For{$i \gets 1$ to $n$}
					\State $\phi_i \gets$   $\{\ell_i(a_{win})\}$ if $i \in$ Fixed, else:  $\ell_i(A)$  
				\EndFor
				\State $A_{iter} \gets \{a \mid \ell_i(a) \in \phi_i, 1 \le i \le n\}$                                                                                    
				\State $a_{win} \gets$ \Call{ConstrainedOneShotNAS}{$A_{iter}$, $\tau_c^{(h)}$}
			\EndFor
			\State \Return $a_{win}$
		\EndFunction
		\State $\mathcal{A}_h \gets$ \Call{HwAwareSearchSpace}{$h$, $\tau_c^{(h)}$} \Comment c.f. Sec~\ref{sec:searchspace}
		\State $a_{win} \gets$ \Call{TwoStageConstrainedNAS}{$\mathcal{A}_h$,  $t$}
		\State retrain $a_{win}$
	\end{algorithmic}
\end{algorithm}

\textbf{Hyper-Parameter $t$.}  The layer grouping boundary $t$ in Algorithm~\ref{alg:ha-nas-2steps} impacts the  effectiveness and efficiency of two-stage acceleration. Specifically, two-stage acceleration  rolls back to the original one-shot NAS and searches for the complete $n$ learnable layers when $t=0$, and thus no search cost reduction is achieved. While larger $t$ reduces the supernet training time a lot, it can harm the search for the optimal architectures. In this paper, we set $t=8$ (only learnable layers counted) according to the natural resolution changes of the supernet. The search space size is reduced by $\approx$65,000 $\times$. According to our empirical results,  two-stage search algorithm achieves
better accuracy and promising search time reduction on two datasets when $t=8$ (c.f. Table~\ref{tbl:task_compare}). We will study more about how to choose $t$ in future works.

%% file: eval.tex
\section{Evaluation}
\label{sec:eval}
\begin{table*}[t]
	\begin{center}
		\small
		\begin{tabular}{l|l|cc|ccccc}
			\hline
			&\multirow{2}{*}{{Model}} &   \multirow{1}{*}{{Search}}&   \multirow{1}{*}{{Target }}  &   \multirow{1}{*}{{FLOPs }}  & \multirow{1}{*}{{DSP }} & \multirow{1}{*}{$\text{  CPU }^\ddagger$} & \multirow{1}{*}{{VPU }}   & \multirow{1}{*}{{Top-1 Acc }} \\
			&&Method&{Hardware}&(\#)&{(ms)}&{(ms)}&{(ms)}&{(\%)}	\\
			\hline
			\multirow{7}{*}{ \rotatebox{90}{\fontsize{7}{7} \selectfont{Existing STOA  Models} }} & 	MobileNetV2~\cite{mobilenetv2} &Manual & CPU  &300M  &10.1    &432.4   &45.2&72.00   \\
			&	MobileNetV3-Large1.0~\cite{mobilenetv3}& RL+NetAdapt~\cite{netadapt}  & CPU   &219M       &141.6    &411.4   &72.3
			&75.20    \\
			
			&	MnasNet-A1~\cite{mnasnet}  & RL  &  CPU   &312M&149.0   &1056.1  &     52.4 &75.20\\
			
			&	FBNet-iPhoneX~\cite{fbnet}    &  Gradient & CPU&322M    &105.0     &313.0   &45.6  &73.20  \\
			&	FBNet-S8~\cite{fbnet}  &  Gradient  &  CPU  &293M   &293.0     &369.6   &45.1   &73.27 \\
			&	Proxyless-R (mobile)~\cite{proxyless} & Gradient & CPU &333M    &534.6   &616.5   &53.1 &74.60       \\

			&	{{\singlepathname}} (block search)~\cite{singlepath} &Oneshot  &-&319M   &270.6     &455.8     &38.7    &74.30   \\
			\hline
			\hline
			
			\multirow{9}{*}{ \rotatebox{90}{\fontsize{7}{7} \selectfont{Search for Our Hardware} }}& 	$\text{Proxyless-R} ^*$& Gradient & DSP &421M     &43.9  &  662.0 & 46.3 &74.20   \\
			&	$\text{{\singlepathname}}^*$   &Oneshot & DSP& 366M &17.3   & 538.2   &   46.1 &74.56    \\
			&		{\sysname}-DSP (Ours)&Oneshot &  DSP &709M    &\textbf{16.5} &576.7    &45.4&\textbf{76.67}   \\ \cline{2-9}
			&	$\text{Proxyless-R} ^*$ & Gradient  &$\text{ CPU }^\ddagger$  &279M    &  182.8 &  392.3 &37.8& 73.40  \\
			&$\text{{\singlepathname}}^*$ &Oneshot &$\text{ CPU }^\ddagger$   &  302M & 106.0  &345.5     & 36.3&  73.76   \\
			
			&		{\sysname}-CPU (Ours)&Oneshot  &$\text{ CPU }^\ddagger$   &327M                &80.1     &\textbf{301.3}  &38.9  &\textbf{74.59}     \\	 \cline{2-9}	 
			&	$\text{Proxyless-R} ^*$ & Gradient  &VPU  &275M & 264.7  & 464.9  &   35.6   &  73.30  \\
			&	$\text{{\singlepathname}}^*$ &Oneshot &VPU&323M &  372.9  &   693.1  &  36.1  & 74.02      \\
			&		{\sysname}-VPU (Ours)&Oneshot   &VPU     &409M             &390.8    &645.3   &\textbf{35.6} &\textbf{75.13}     \\
			\hline
		\end{tabular}
	\end{center}
	\caption{Compared with state-of-the-art hardware-aware NAS methods on ImageNet, {\sysname} 
		is the only NAS method that consistently achieves high accuracy and low latency on all the target hardware.  Latency numbers are measured on our hardware platforms. $^\star$: We run Proxyless-R and SPOS with their officially open-sourced implementations to search models on our hardware platforms .   $\ddagger$: CPU latency is measured on a single CPU core with float32 precision.   }
	\label{tbl:imagenet}

\end{table*}

\begin{table}[h]
		\begin{center}
		\small
		\begin{tabular}	{c|c|c|c}
			\hline
		\textbf{Group}	&\textbf{NAS} &\textbf{Acc} & \textbf{CPU} \\
				&&\textbf{(\%)} & \textbf{(ms)} \\
			\hline
			
	\multirow{3}{*}{{Similar latency}}	&	FBNet-iPhoneX & 73.20&313.0\\
		&	FBNet-S8 &73.27&369.6\\
		&	{\sysname}-CPU& \textbf{74.59}&301.3\\
			\hline          
	\multirow{3}{*}{Similar accuracy}	&	Proxyless-mobile&74.60 & 616.5\\
	&FBNet-C&74.90&688.6\\
		&	MnasNet-A1&75.20&1056.1\\
		&	{\sysname}-CPU1 & 74.98&\textbf{381.2}\\
		\hline
		\end{tabular}
	\end{center}
	\caption{Compared with models of same-level CPU inference latency, {\sysname}-CPU improves the top-1 accuracy from 73.27\% to 74.59\% on ImageNet. Compared with models of same-level top-1 accuracy, {\sysname}-CPU1 accelerates the CPU inference time by 1.62$\times$ - 2.77$\times$.  }
	\label{tbl:compare_cpu}
\end{table}

			

\subsection{Experiment Setup}

\textbf{Hardware Platforms and Measurements}. We target three representative mobile hardware that is widely used for CNN deployment: (1) DSP (Qualcomm's $\text{Hexagon}^\text{TM}$ 685 DSP),  (2): CPU (Qualcomm's Snapdragon 845 ARM CPU), (3): VPU (Intel's $\text{Movidius}^\text{TM}$ $\text{Myriad}^\text{TM}$ X Vision Processing Unit ). To make full utilization of these hardware at inference, we use the particular inference engine provided by the hardware vendor. Specifically, DSP and CPU latency are measured by Snapdragon Neural Processing Engine SDK~\cite{snpe} with int8 and float32 precision, respectively.   VPU latency is measured by Intel $\text{OpenVINO}^{\text{TM}}$ Toolkit~\cite{movidius} with float16 implementation. 

\textbf{Latency Constraints}. For better comparison with other works, we set the latency constraints to be smaller than the best latency of models from other works~\cite{mnasnet,mobilenetv3,fbnet,proxyless}, which are 310 $ms$ (CPU), 17 $ms$ (DSP) and 36 $ms$ (VPU).

\textbf{Hardware-aware One-Shot Search}. As shown in Table~\ref{tbl:searchspace}, the hardware aware search space is generated according to the different characteristics of every hardware.  
The search space is then encoded into a over-parameterized supernet for two-stage acceleration search. 
 Our two-stage search acceleration is built on top of {\singlepathname}~\cite{singlepath,singlepath_code}. Once the supernet training finishes, we perform a 20-iterations evolution search for total 1,000 architectures as {\singlepathname}. To avoid measuring the latency of every candidate architecture during search, we build a latency-prediction model \footnote{ The prediction model is built with Bayesian Ridge Regression~\cite{bayesianregression}} with high accuracy: the average estimated error for DSP, CPU, VPU is 4.7\%, 4.2\%, and 0.08\%, respectively.

\subsection{Searching on ImageNet Dataset}

Our comparisons are two-folds: (1) we compare {\sysname} searched models with various existing state-of-the-art efficient models that are primarily designed or searched for ARM CPU, to demonstrate that {\sysname} is able to generate different models suitable for different types of hardware; (2) we compare {\sysname} with two representative NAS methods, Proxyless-R~\cite{proxyless} and SPOS~\cite{singlepath} \footnote{Considering the reproducibility issue, we didn't test other hardware-aware NAS methods due to the lack of officially open-sourced code.}, to show that {\sysname} is able to search for better models at a lower cost, benefiting from the two-stage search algorithm.  The primary metrics we care about are top-1 \textbf{accuracy} on the ImageNet dataset and \textbf{inference latency} on the three hardware.
 
\textbf{Dataset and Training Details.} Following~\cite{proxyless}, we randomly split the original training set into two parts: 50,000 images for validation (50 images for each class exactly) and the rest as the training set. The original validation set is used for testing, on which all the evaluation results are reported. We follow most of the training settings and hyper-parameters used in {\singlepathname}~\cite{singlepath}, with two exceptions: \textit{(i)} For supernet training, the  epochs change with different hardware-aware search spaces (listed in Table~\ref{tbl:searchspace}), and we stop at the same level training loss as {\singlepathname}. \textit{(ii)} For architecture retraining, we change linear learning rate decay to cosine decay from 0.4 to 0. The batch size is 1,024. Training uses 4 \textit{NVIDIA V100} GPUs. We implement  Proxyless-R~\cite{proxyless_code} and SPOS~\cite{singlepath_code} on our hardware platforms to search for the  models within the same latency constraints.

		
		

\textbf{Results and Analysis.} 
 Table~\ref{tbl:imagenet} and Table ~\ref{tbl:compare_cpu} summarize our experiment results on ImageNet. 
  It demonstrates that it's essential to leverage hardware diversity in NAS to consistently achieve the high accuracy and low latency on different hardware platforms.  
 
  Firstly, {\sysname} surpasses existing  state-of-the-art efficient models. Compared to MobileNetV2 (top-1 accuracy 72.0\%), {\sysname} improves the accuracy by 2.59\% to 4.03\% on all target hardware platforms. Compared to state-of-the-art models searched by NAS, {\sysname} achieves the lowest inference latency on DSP, CPU, VPU, with better or comparable accuracy. Remarkably, {\sysname}-DSP achieves 76.67\% accuracy, better than MnasNet-A1 (+1.47\%), FBNet-iPhoneX (+3.47\%), FBNet-S8 (+3.4\%), Proxyless-R (+2.07\%), and {\singlepathname} (+2.37\%). Regarding latency, {\sysname}-DSP is 16.5$ms$ on DSP, that reaches a  a 6.35$\times$ inference speedup than FBNet-iPhoneX. Interestingly,
{\sysname}-DSP is faster than other NAS models but with a much larger FLOPs count. This is against the widely accepted belief that smaller FLOPs count results in lower latency. Our study for DSP indicates that small-kernel-sized complicated operators are most suitable for this platform, and the hardware aware search space fully takes advantage of this and benefits from such operators (c.f. Sec~\ref{sec:searchspace}). 

Secondly, {\sysname} outperforms the state-of-the-art  NAS methods for the same target hardware platforms. Compared to SPOS, {\sysname} improves the accuracy by  2.11\% (DSP), 0.83\% (CPU), 1.11\% (VPU) with slightly lower inference latency (DSP: -0.8ms, CPU: -44.2ms, VPU: -0.5ms). When compared with Proxyless-R,  our method achieves higher accuracy of 2.47\% (DSP), 1.49\% (CPU), 1.83\% (VPU) with less inference latency (DSP: -27.4ms, CPU: -91ms). We noted that the models searched by  Proxyless-R are with lower accuracy and larger latency than SPOS and ours.  One hypothesis is that the default hyper-parameter $w$ in Proxyless-R that controls the trade-off between accuracy and latency might be not optimal for other hardware platforms. For instance, Proxyless-R searched many zero operators for the earlier layers on CPU and VPU.

Finally, {\sysname} also achieves competitive performance on the well-studied ARM CPU. To further compare the efficiency on CPU, we group related NAS models into same-level latency group and same-level accuracy group in Table~\ref{tbl:compare_cpu}. For fairness, we didn't compare MobileNetV3-Large1.0 as it adopts a second fine-grained search by NetAdapt on MnasNet-A1. Results in Table~\ref{tbl:compare_cpu} suggests {\sysname} (CPU) achieves the highest accuracy in same-level CPU inference time group, and achieves 1.62$\times$ - 2.77$\times$ lower inference time in same-level accuracy group.

\textbf{Search Cost Analysis}. To compare the search cost, we report supernet training time reduction compared with {\singlepathname} instead of exact GPU search days as ~\cite{proxyless,fbnet} for two reasons: \textit{(i)}: the GPU search days are highly relevant with the experiment environments (e.g., different GPU hardware) and the code implementation (e.g., ImageNet distributed training). \textit{(ii)}: The primary time cost comes from supernet training in {\singlepathname}, as the evolution search is fast that architectures only perform inference.  

Compared with {\singlepathname}, {\sysname} (\textit{Stage1 + Stage2}) reduces 30.4\% supernet training time and finds models with better performance. Furthermore, {\sysname} (\textit{Stage1}) already achieves better classification accuracy than other NAS methods (DSP: 76.57\%, CPU: 74.59\%, VPU: 74.63\%) while reducing an average of 54.7\% time, which is almost a 2$\times$ training time speedup (More analysis are in Sec~\ref{two-stage}). It demonstrates the effectiveness of exploring more architecture  in the deeper CNN layers.

\subsection{Ablation Study and Analysis}
\label{sec:compare_searchmethod}

\begin{table*}[t]
	\begin{center}  
		\small
		\begin{tabular}	{l|c|cc|cc|cc}
			\hline
			\multirow{4}{*}{\textbf{Search Space}}&	\multirow{4}{*}{\textbf{Hardware}}&\multicolumn{2}{c|}{\textbf{{\singlepathname}}} &\multicolumn{2}{c|}{\textbf{{\sysname}}} &\multicolumn{2}{c}{\textbf{{\sysname}}}\\
			&& \multicolumn{2}{c}{ \textit{($20$ $layers$)}} &\multicolumn{2}{c|}{ \textit{($Stage1$: 12 layers)}}    &\multicolumn{2}{c}{\textit{($Stage1+Stage2$: 20 layers)}}\\
			&	&\hspace{-4px} \textbf{Acc }                      &\hspace{-4px}\textbf{Train}       &\hspace{-4px} \textbf{Acc} &\hspace{-4px}\textbf{Train }&\hspace{-4px} \textbf{Acc } &\hspace{-4px}\textbf{Train }    \\
			&	&\textbf{(\%)}   &\textbf{iters (\#)}   &\textbf{(\%)}  &\textbf{iters (\#)}  &\textbf{(\%)}  &\textbf{iters (\#)}\\
			
			\hline	
				\multirow{3}{*}{{Manually-designed}}&DSP  &86.35&\hspace{-4px}393,000 & 86.29&\hspace{-4px} 157,200& 86.69&\hspace{-4px} 184,710 \\

			&CPU &86.52&\hspace{-4px}393,000 &86.32 &\hspace{-4px}  157,200&86.75 &\hspace{-4px} 183,400 \\
				&VPU &86.68&\hspace{-4px} 393,000& 86.49&\hspace{-4px} 157,200 &86.90 &\hspace{-4px} 192,000 \\
			\hline
			\multirow{3}{*}{Hardware-aware}&DSP  &87.22 &\hspace{-4px}569,850 &86.56 &\hspace{-4px}128,380 &87.62 &\hspace{-4px}150,650  \\
			& CPU&87.02 &\hspace{-4px}476,840 &86.75 &\hspace{-4px}144,100 &87.33 & \hspace{-4px}168,990   \\
	
		
			&VPU&86.99 &\hspace{-4px}524,000 &86.93 &\hspace{-4px}133,620 &87.07 & \hspace{-4px}157,200   \\
			\hline
		\end{tabular}
	\end{center}
	\caption{Compared to {\singlepathname}~\cite{singlepath}, our two-stage search method achieves higher accuracy with much less search cost (51.1\%-70\%) on both manually-designed and hardware-aware search spaces. We list out the training iterations OUI (batchsize=64) for search cost comparison.}
	\label{tbl:task_compare}
\end{table*}

We now further evaluate the efficiency of the proposed hardware-aware search space and two-stage acceleration algorithm. For simplicity, we run the experiments on the OUI-Adience-Age (OUI) dataset. OUI~\cite{oui-adience-age} is a  small 8-class dataset consisting of  17,000 face images. We split OUI into train and test sets by 8:2. 
The settings for supernet training  parameters are the same as ImageNet experiments except that we reduce the initial learning rate from 0.5 to 0.1, and the batch size reduced from 1024 to 64. Supernet trains until converge. For the architecture retraining, we train for 400 epochs and change the linear learning rate decay to 
Cyclic decay~\cite{cyclic} with a $[0,1]$ bound.  
We use 1 \textit{NVIDIA Telsa P100} for training.
\subsubsection{Effectiveness of Hardware-aware Search Space}
\label{sec:hw-insights}

\begin{figure}[t]
	\centering
	\includegraphics[width=0.9\columnwidth]{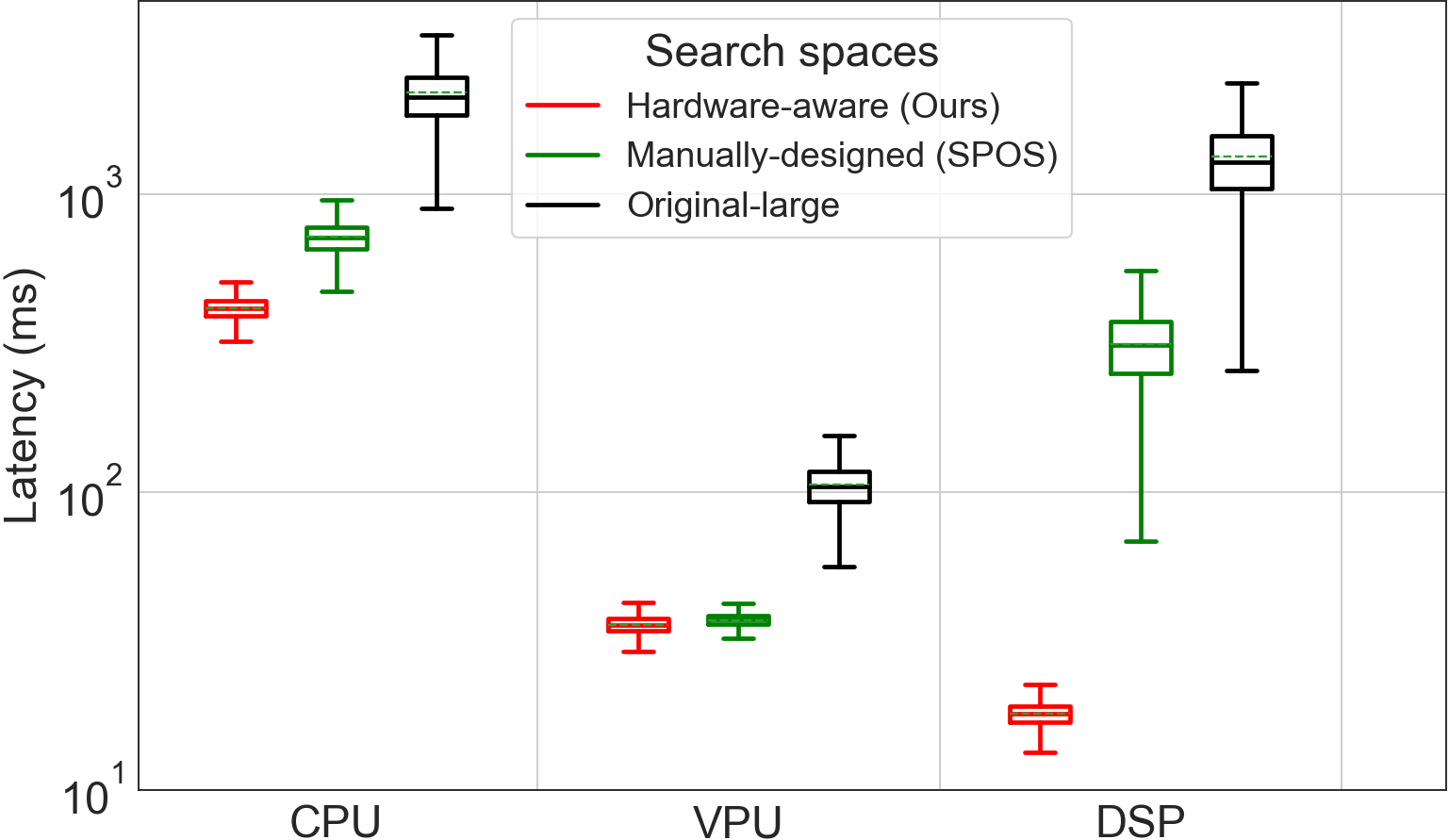}
	
	\caption{ Architectures sampled in hardware specialized search spaces achieve lower latency than manually-designed and original-large search spaces. The y-axis is log-scaled for better comparison.  }
	\label{fig:searchspacecompare}
\end{figure}
\begin{figure}[t]
	\centering
	\includegraphics[width=0.92\columnwidth]{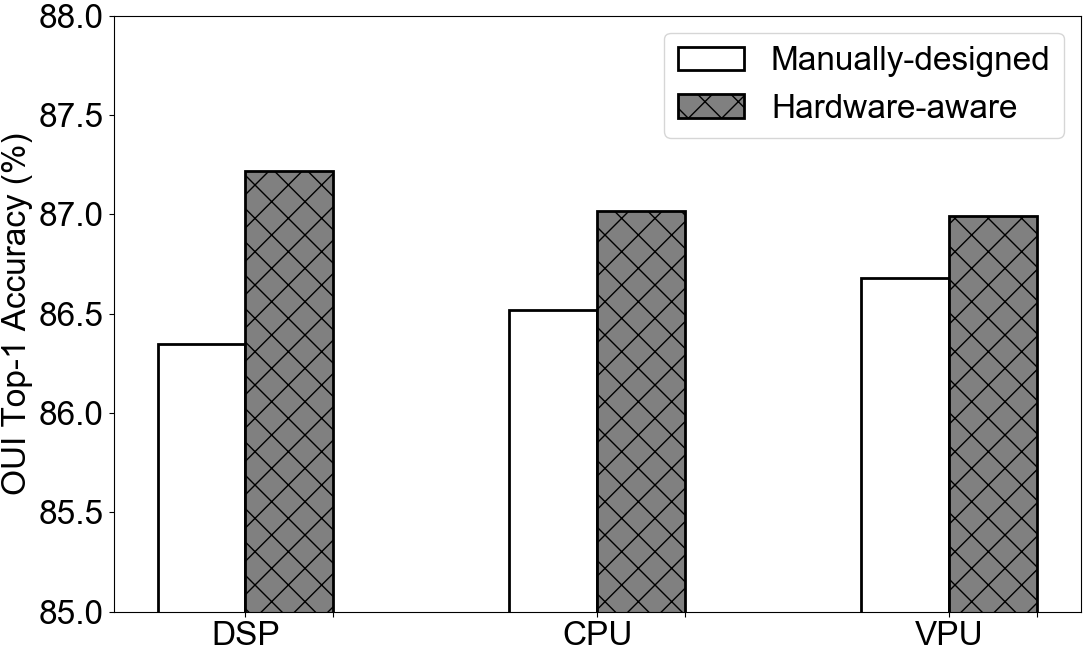}
	\caption{ Compared to manually-designed search space, our hardware-aware search spaces achieve higher accuracy by {\singlepathname}.}
	\label{fig:sp_acc}
\end{figure}

Ideal search spaces contain many high accuracy architectures within the latency constraint so that the optimal ones are easily sampled by the search algorithm. We first investigate the latency distribution of architectures in a search space. We random sample 10 million architectures from three different search spaces: \textit{(i)}: our hardware-aware search spaces (c.f. Table~\ref{tbl:searchspace}), \textit{(ii) Manually-designed}: the search space used in {\singlepathname}~\cite{singlepath} that designed by domain experts, \textit{(iii) Original-large}: the search space generated by our large operator pool (c.f. Table~\ref{tbl:operators}) , and benchmark their inference latency on hardware. As Figure~\ref{fig:searchspacecompare} shows, architectures in our hardware specialized search spaces are with much lower  latency than the manually-designed and the original large search spaces. This indicates our search space is more compact and easier for constrained sampling. 

To evaluate the accuracy gains by search space, we run the  {\singlepathname} on both manually-designed and hardware-aware search spaces. We did not compare with the original-large search space here due to the unacceptable cost and  non-convergence problem in supernet training. 
Figure~\ref{fig:sp_acc} shows our hardware-aware search space consistently achieves higher  accuracy than the manually-designed search space ( +0.87\% on DSP, +0.5\% on CPU, +0.31\% on VPU). 
 
\textbf{Hardware Insights}. 
We  share several important \textit{insights} from  search space generation.
\begin{itemize}
	\item \textit{$\text{Hexagon}^\text{TM}$ 685 DSP}. Small kernel convolutions ($k\le3$) are well optimized on this platform. As a result, all the operators are of $k$=3 in search space. It also allows the search space to contain complicated operators (of large FLOPs) with small kernels, because their efficiency on this platform is better than those less-complicated operators but with bigger kernels. That's why  {\sysname} (DSP) is faster than other NAS models but with a much larger FLOPs. On the contrary, the search spaces of Proxyless and {\singlepathname} contain many large kernel operators (i.e., $k$=5/7).
	\vspace{-4px}
	\item \textit{$\text{Myriad}^\text{TM}$ X VPU}. The efficiency is strongly impacted by whether the operator is natively supported by the AI accelerator. For example, $SE$ module is of low efficiency in this platform, because it has to roll back to relatively slow CPU execution. On the contrary, convolutions with bigger kernels ($k=7$) are much more efficiently executed than on other platforms. This explains why the search space for this platform selects no $SE$ operators but much more bigger kernel operators (especially in the earlier layers). We also observe that the operators applied in  Proxyless and {\singlepathname} are all supported by the AI accelerator. Therefore, the VPU latency of searched models by Proxyless-R and {\singlepathname} are relatively low as shown in Table~\ref{tbl:imagenet}. 
 
	\vspace{-4px}
	\item \textit{Snapdragon 845 ARM CPU}. Even with complex memory operator, Choice\_3 (i.e., ShuffleNetV2 unit) is the most efficient operator on this platform.
\end{itemize}

\subsubsection{Effectiveness of Two-Stage Search Algorithm}
\label{two-stage}
Different with previous NAS methods that globally search over all the learnable layers (e.g., SPOS searches $n=20$ layers), our two-stage search algorithm groups CNN layers into 12 later and 8 earlier layers: Stage1 searches the later  layers first, Stage2 searches the earlier layers for the winning architectures in Stage1.   To demonstrate the effectiveness, we compare it with {\singlepathname} on both manually-designed and our hardware-aware search spaces.



 Table~\ref{tbl:task_compare} summarizes experiment results.
 On all search spaces, our proposed method could achieve not only higher accuracy but also less search cost for the target hardware under the latency constraint. In addition, only one step search (\textit{Stage1}) of {\sysname} could achieve a comparable top-1 accuracy (with an average of 0.23\% loss), but the number of training iterations is significantly reduced (60\%-77.5\%). This indicates that operators in later CNN layers are more critical for final accuracy. 
 
If the computation budget (e.g., training time) allows, {\sysname} can further benefit from the second step search (\textit{Stage2}). The accuracy is improved by 0.14\%-1.06\% with an additional cost of only 4.0\%-8.8\% of training iterations. Our gains mainly come from the reduced search space size by the two-stage search algorithm.

%% file: conclusion.tex
\section{Conclusion}
In this paper, we propose {\sysname} to address the challenge of hardware diversity in NAS. By exploring hardware-aware search space and two-stage search algorithm, we demonstrate that {\sysname} is able to search for better models specialized for different hardware platforms and outperforms the previous NAS methods by both accuracy and significant training time reduction. 


%% file: models.tex
\newpage

\begin{figure*}
	\renewcommand\thefigure{4}    

	\begin{center}  
	
	\subfigure[Choice\_k ]{
		\includegraphics[width=0.18\linewidth]{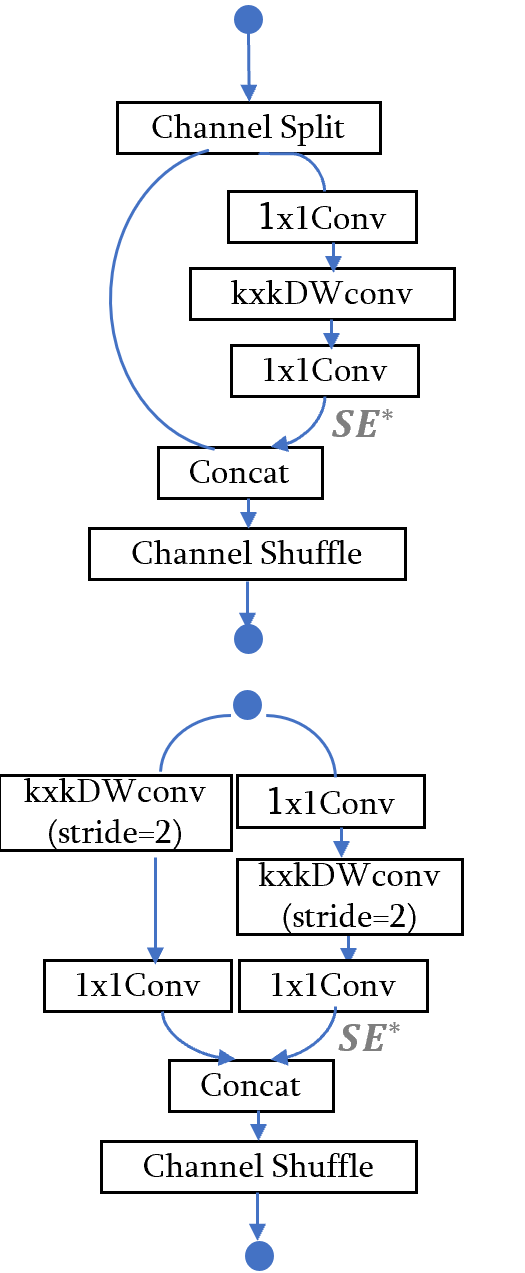}
		\label{fig:op_choicek}
		
	}
	\subfigure[ChoiceX ]{
		\includegraphics[width=0.16\linewidth]{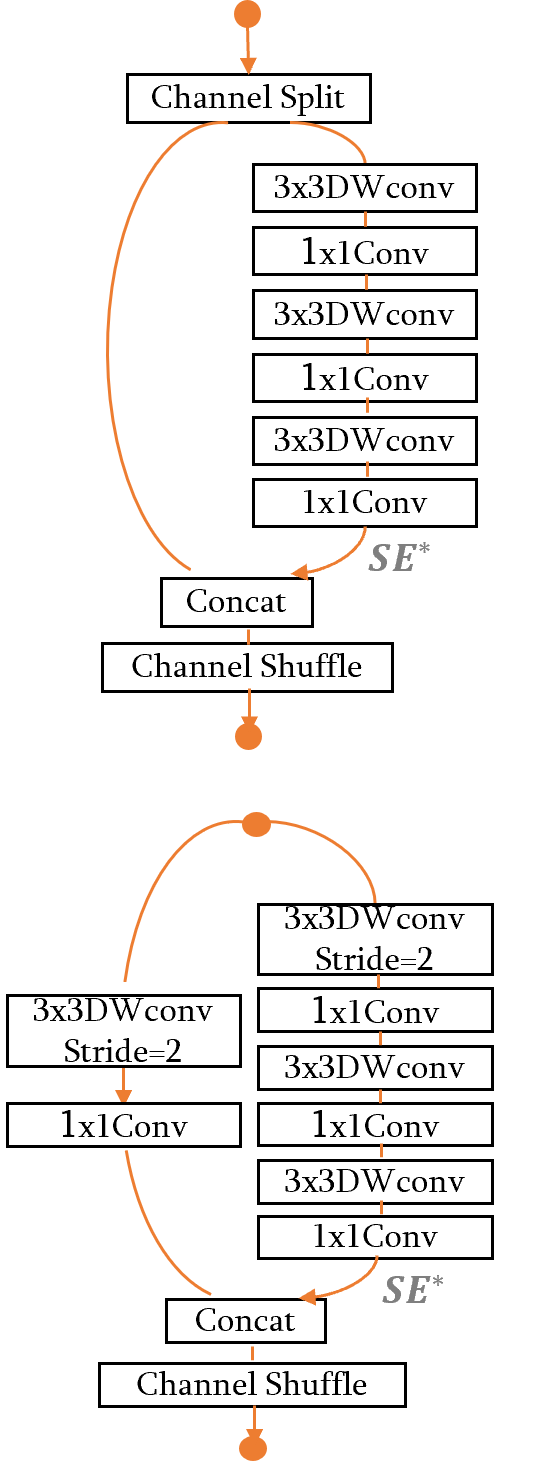}
		\label{fig:op_choicex}
		
	}
	\subfigure[SEP\_k]{
		\includegraphics[width=0.15\linewidth]{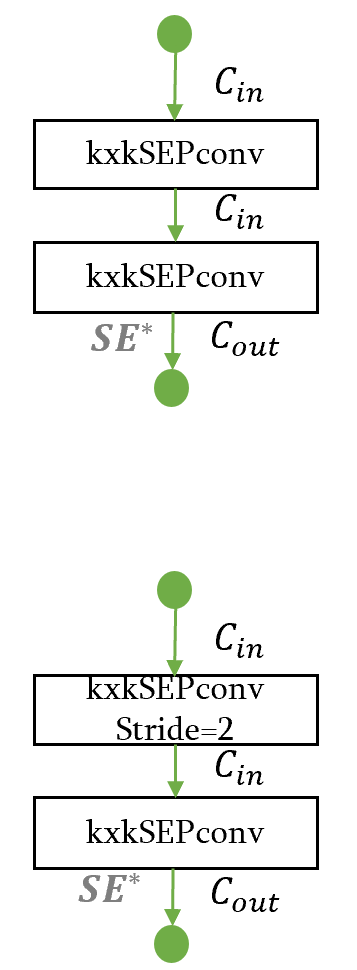}
		\label{fig:op_sep}
		
	}
	\subfigure[MB\_k\_e]{
		\includegraphics[width=0.15\linewidth]{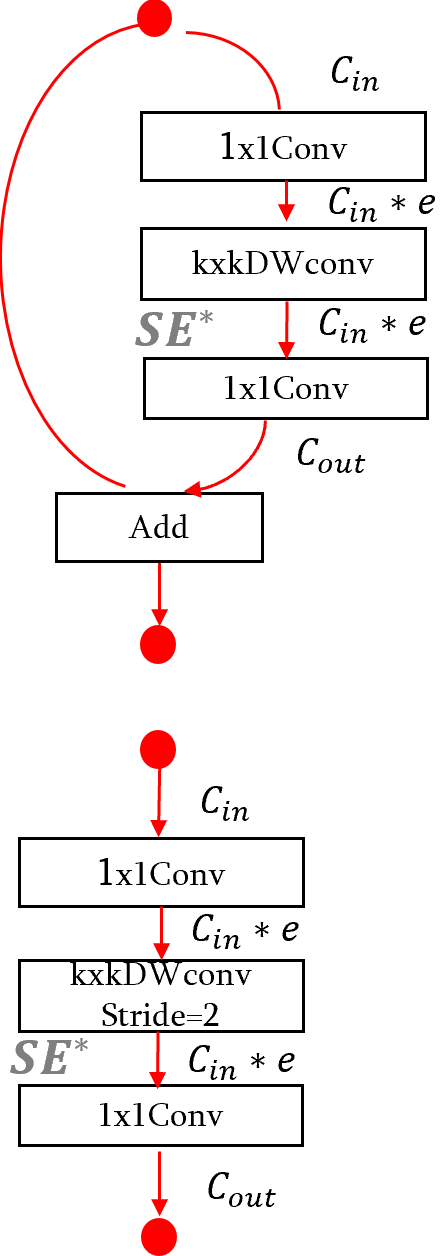}
		\label{fig:op_mb}
		
	}
	\end{center}
	\caption{Operators in Table 1. $SE^*$ indicates the position to add squeeze-and-excitation block. When $SE^*$ is enabled, the operations are Choice\_k\_SE, ChoiceX\_SE, SEP\_k\_SE, MB\_k\_e\_SE.  $k$ indicates kernel size, where $k=3,5,7$. $e$ indicates the expansion rate, where $e=1,3,6$. }
	\label{fig:operations}
\end{figure*}

\begin{figure*}
	\renewcommand\thefigure{5}  
	\begin{center}  

	\includegraphics[width=0.87\textwidth]{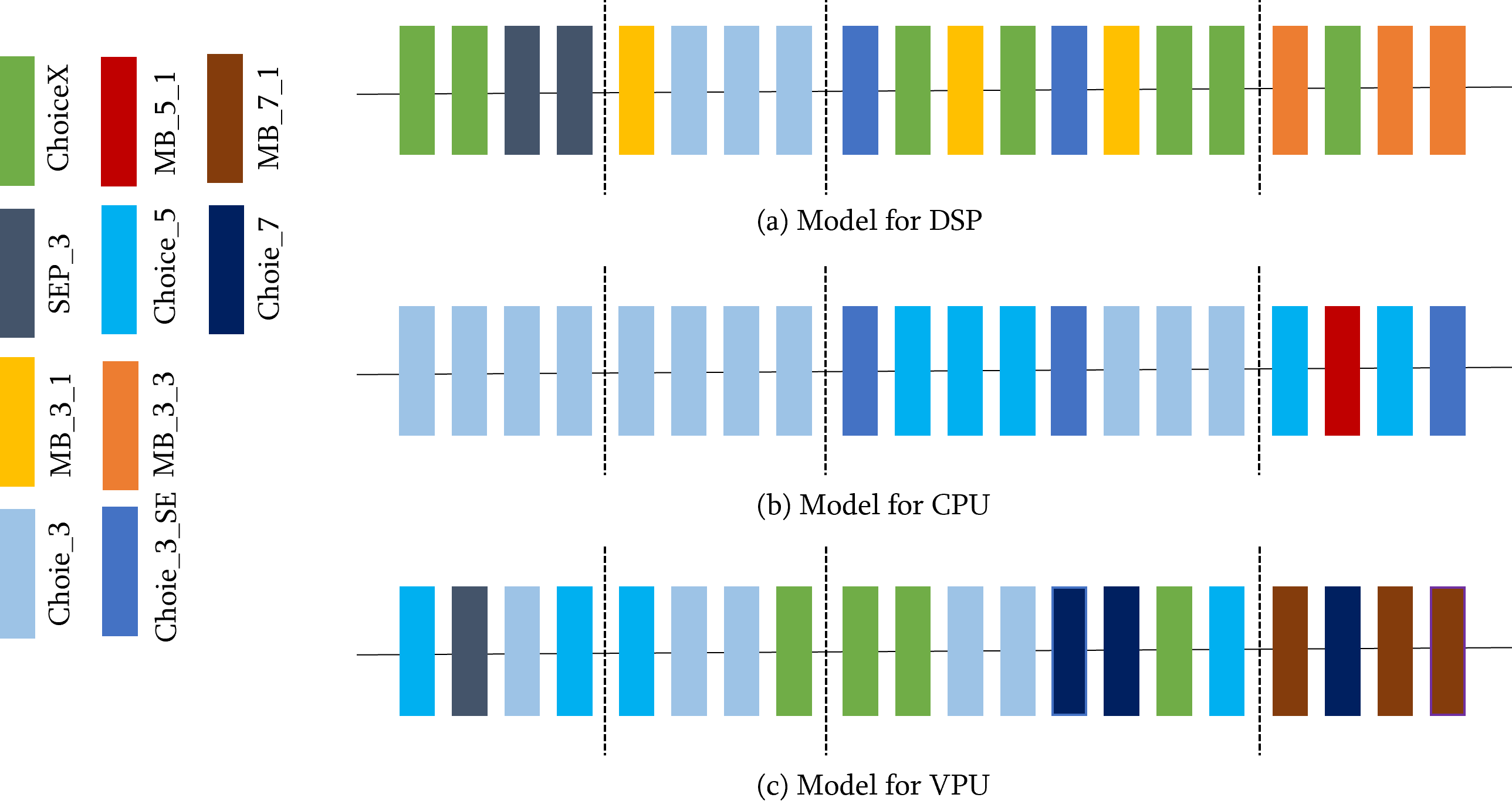}
	\end{center}
	
	\caption{Structures of searched architectures on ImageNet in Table 3. }
	\label{fig:bestmodels}
\end{figure*}